\pdfoutput=1
\documentclass[11pt]{article}
\usepackage{graphicx}
\usepackage[ruled,vlined]{algorithm2e}
\usepackage{graphicx}
\usepackage[]{emnlp2021}
\usepackage{comment}
\usepackage{times}
\usepackage{latexsym}
\newcommand{\mehdi}[1]{\textcolor{cyan}{#1}}

\usepackage[T1]{fontenc}

\usepackage[utf8]{inputenc}

\usepackage{microtype}
\usepackage{amsmath}
\usepackage{ulem}
\usepackage{enumitem}
\title{KroneckerBERT: Learning Kronecker Decomposition for Pre-trained Language Models via Knowledge Distillation}

\author{Marzieh S. Tahaei \\
    Noah’s Ark Lab,\\
    Huawei Technologies Canada\\
     {marzieh.tahaei@huawei.com} \\\And
    Ella Charlaix \\
    Noah’s Ark Lab,\\
    Huawei Technologies Canada\\
    {charlaixe@gmail.com}\\\And
     Vahid Partovi Nia \\
    Noah’s Ark Lab\\
    Huawei Technologies Canada\\
    {vahid.partovinia@huawei.com} \\\AND
    Ali Ghodsi\\
    Department of Statistics and \\
    Actuarial Science, University of Waterloo\\
    {ali.ghodsi@uwaterloo.com}\\\And
     Mehdi Rezagholizadeh\\
         Noah’s Ark Lab,\\
    Huawei Technologies Canada\\
    {mehdi.rezagholizadeh@huawei.com}}

\begin{document}
\maketitle
\begin{abstract}
The development of over-parameterized pre-trained language models has made a significant contribution toward the success of natural language processing. While over-parameterization of these models is the key to their generalization power, it makes them unsuitable for deployment on low-capacity devices. We push the limits of state-of-the-art Transformer-based pre-trained language model compression using Kronecker decomposition. We use this decomposition for compression of the embedding layer, all linear mappings in the multi-head attention, and the feed-forward network modules in the Transformer layer. We perform intermediate-layer knowledge distillation using the uncompressed model as the teacher to improve the performance of the compressed model. We present our KroneckerBERT, a compressed version of the $\text{BERT}_\text{BASE}$ model obtained using this framework. 
We evaluate the performance of KroneckerBERT on well-known NLP benchmarks and show that for a high compression factor of 19 (5\% of the size of the $\text{BERT}_\text{BASE}$ model), our KroneckerBERT outperforms state-of-the-art compression methods on the GLUE. Our experiments indicate that the proposed model has promising out-of-distribution robustness and is superior to the state-of-the-art compression methods on SQuAD. 
\end{abstract}

\section{Introduction}
\raggedbottom
In recent years, the emergence of \textit{Pre-trained Language Models} (PLMs) has led to a significant breakthrough in Natural Language Processing (NLP).
The introduction of Transformers and unsupervised pre-training on enormous unlabeled data are the two main factors that contribute to this success.

Transformer-based models \citep{devlin2018bert,radford2019language,yang2019xlnet,shoeybi2019megatron} are powerful yet highly over-parameterized. The enormous size of these models does not meet the constraints imposed by edge devices on memory, latency, and energy consumption. Therefore there has been a growing interest in developing new methodologies and frameworks for the compression of these large PLMs. Similar to other deep learning models, the main directions for the compression of these models include low-bit quantization \citep{gong2014compressing, prato2019fully}, network pruning \citep{han2015deep}, matrix decomposition \citep{yu2017compressing, lioutas2020improving} and Knowledge distillation \citep{hinton2015distilling}. These methods are either used in isolation or in combination to improve compression-performance trade-off.

Recent works \citep{sanh2019distilbert,sun2019patient,jiao2019tinybert,sun2020mobilebert,xu2020bertoftheseus} have been quite successful in compressing Transformer-based PLMs to a certain degree; however, extreme compression of these model (compression factors >10) is still quite challenging. Several works \citep{mao2020ladabert,zhao2019extreme,zhao2021extremely} that have tried to go beyond the compression factor of 10, have done so at the expense of a significant drop in performance. 
This work proposes a novel framework that uses Kronecker decomposition for extreme compression of Transformer-based PLMs and provides a very promising compression-performance trade-off.  Similar to other decomposition methods, Kronecker decomposition can be used to represent weight matrices in NNs to reduce the model size and the computation overhead. Unlike the well-known SVD decomposition, Kronecker decomposition retains the rank of the matrix and hence provides a different expressiveness compared to SVD. 

We use Kronecker decomposition for the compression of both Transformer layers and the embedding layer. For Transformer layers, the compression is achieved by representing every weight matrix both in the multi-head attention (MHA)  and the feed-forward neural network (FFN) as a Kronecker product of two smaller matrices. We also propose a Kronecker decomposition for compression of the embedding layer.  Previous works have tried different techniques to reduce the enormous memory consumption of this layer \citep{khrulkov2019tensorized,li2018slim}. Our Kronecker decomposition method can substantially reduce the amount of required memory while maintaining low computation.

Using Kronecker decomposition for large compression factors, which is the focus of this paper, leads to a reduction in the model expressiveness. 
To address this issue, we propose to distill knowledge from the intermediate layers of the original uncompressed network to the Kronecker network during training. 
We use this approach for compression of the $\text{BERT}_\text{BASE}$ model.
We show that for a large compression factor of 19$\times$, KroneckerBERT provides significantly better performance than state-of-the-art compressed BERT models. While our evaluations in this work are limited to  BERT, this framework can be easily used to compress other Transformer-based NLP models. The main contributions of this paper are as follows:
\begin{itemize}[noitemsep]
    \item Developing a framework for compression of the embedding layer using the Kronecker decomposition.
    \item Deploying the Kronecker decomposition for the compression of Transformer modules.
    \item Proposing a KD method to improve the performance of the Kronecker network.
    \item Evaluating the proposed framework for compression of BERT$_{\text{BASE}}$ model on well-known NLP benchmarks
\end{itemize}
\begin{figure}
    \centering
    
    \includegraphics[scale=0.085]{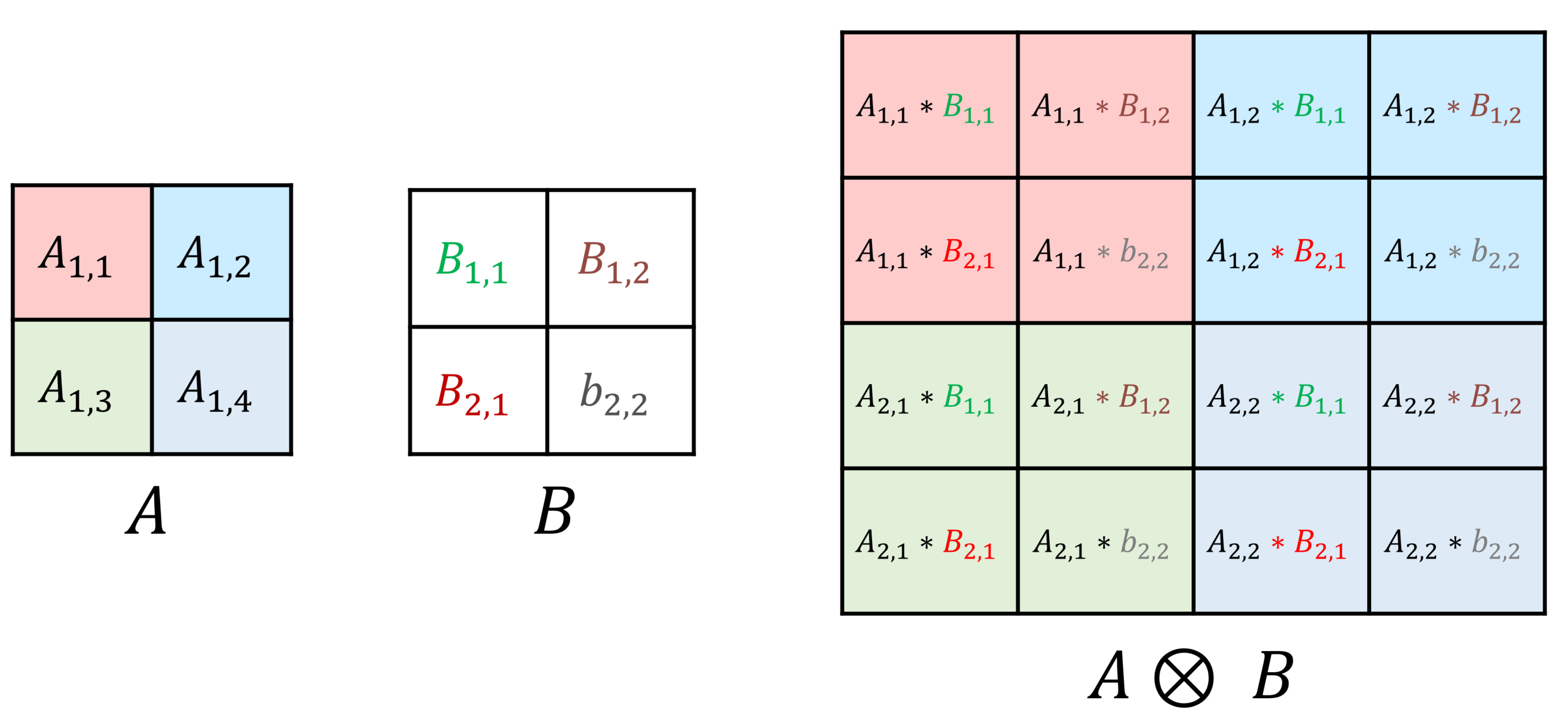}
    
    \caption{An example of Kronecker product of two 2 by 2 matrices}
    \label{fig:kronecker_product}
\end{figure}

\section{Related Work}
In this section, we first go through some of the most related works for BERT compression in the literature and then review the few works that have used Kronecker decomposition for compression of CNNs and RNNs. 
\subsection{Pre-trained Language Model Compression}
In recent years, many model compression methods have been proposed to reduce the size of PLMs while maintaining their performance on different tasks. KD, which was first introduced by \citep{bucilu2006model}  and then later generalized by \citep{hinton2015distilling}, is a  popular compression method where a small student network is trained to mimic the behavior of a larger teacher network. 
Recently, using KD for the compression of PLMs has gained a growing interest in the NLP community. BERT-PKD \citep{sun2019patient}, uses KD to transfer knowledge from the teacher's intermediate layers to the student in the fine-tuning stage. TinyBERT \citep{jiao2019tinybert} uses a two-step distillation method applied both at the pre-training and at the fine-tuning stage. MobileBERT \citep{sun2020mobilebert} also uses an intermediate-layer knowledge distillation methodology, but the teacher and the student are designed by incorporating inverted-bottleneck. Several works combine different compression techniques such as knowledge distillation and pruning with matrix factorization \citep{mao2020ladabert} or quantization \citep{kim2020fastformers}. In \citep{mao2020ladabert}, the authors present LadaBERT, a lightweight model compression pipeline combining SVD-based matrix factorization with weight pruning, as well as a knowledge distillation methodology as described in \citep{jiao2019tinybert}.

To achieve higher compression factors, authors in  \citep{zhao2019extreme} use a layer-wise KD method to reduce the size of vocabulary (from the usual 30k to 5k) and the width of the layers. Similarly, \citealt{zhao2021extremely} uses a mixed-vocabulary training method to train models with a smaller vocabulary.

\subsection{Kronecker Factorization}
Kronecker products have previously been utilized for the compression of CNNs and small RNNs. \citealt{zhou2015compression} was the first work that utilized Kronecker decomposition for NN compression. They used summation of multiple  Kronecker products to replace weight matrices in the fully connected and convolution layers in simple CNN architectures like AlexNet. \citealp{thakker2020compressing} used Kronecker product for the compression of very small language models for deployment on IoT devices. To reduce the amount of performance drop after compression, they propose a hybrid approach where the weight matrix is decomposed into an upper part and lower part. The upper part remains un-factorized, and only the lower part is factorized using the Kronecker product. More recently, \citealt{thakker2020compressing} tried to extend the previous work to non-IoT applications. Inspired by robust PCA, they add a sparse matrix to Kronecker product factorization and propose an algorithm for learning these two matrices together.

To the best of our knowledge, this work is the first attempt to compress Transformer-based language models using Kronecker decomposition. Unlike prior arts, this work uses a simple Kronecker product of two matrices for the representation of linear layers and uses KD framework to improve the performance of the compressed model.

\begin{figure}
    \centering
    \includegraphics[scale=0.62]{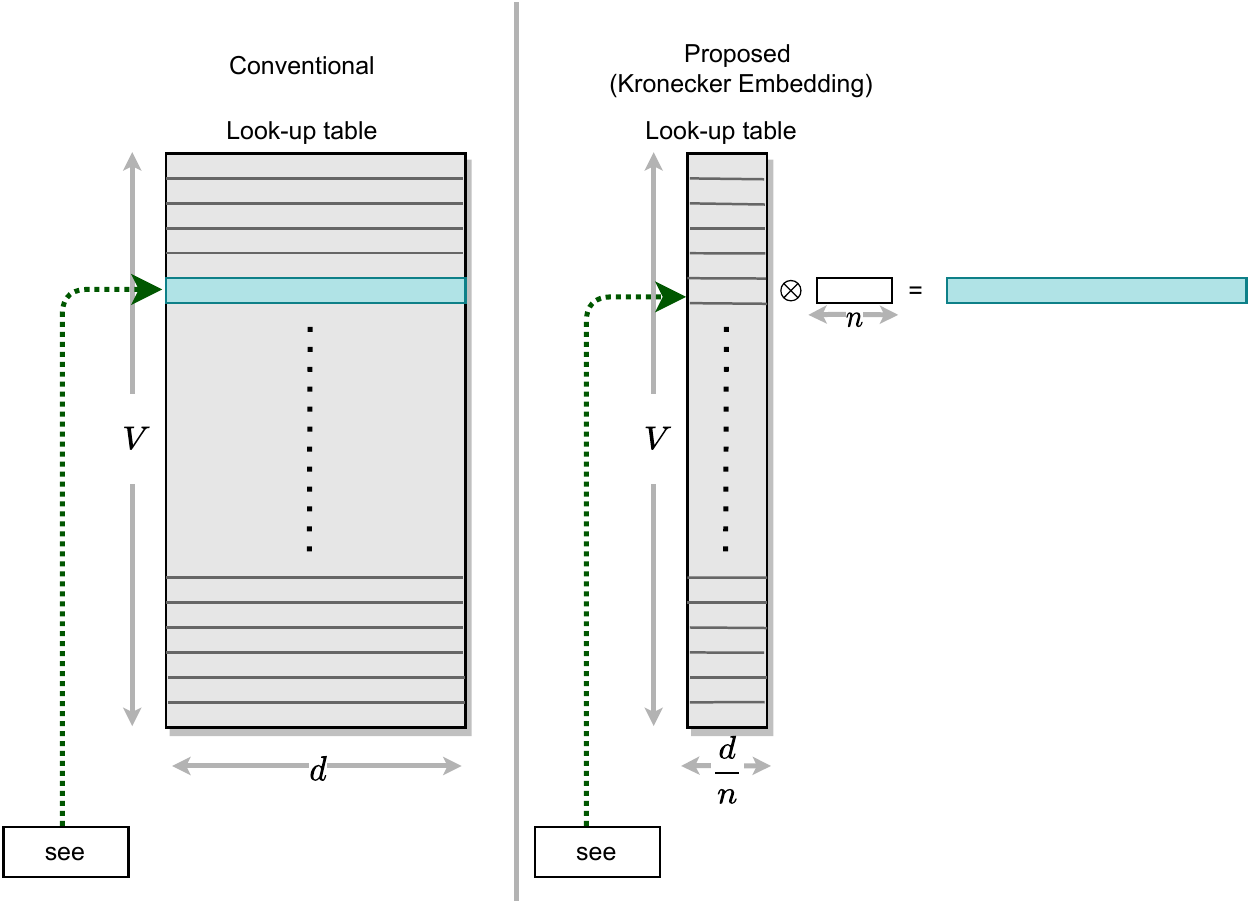}
    \caption{Illustration of our proposed method for the compression of the embedding layer. Left: conventional embedding stored in a lookup table. Right: Our proposed compression method where the original embedding matrix is represented as a Kronecker product of a matrix and a row vector. The matrix is stored in a lookup table to minimize computation over head.}
    \label{fig:embedding}
\end{figure}

\section{Methodology}
In this section, we first introduce the background of Kronecker decomposition. We then explain our compression method in detail.
\subsection{Kronecker Product}
Kronecker product is an operation that is applied on two matrices resulting in a block matrix. Let $A$ be a matrix $\in R^{m_1\times n_1}$, and let $B$ be a matrix $\in R^{m_2\times n_2}$, then the Kronecker product of $A$ and $B$ denoted by $\otimes$ is a block matrix, where each block $(i,j)$ is obtained by multiplying the element $A_{i,j}$ by matrix $B$. Therefore, the of the resulting matrix $A\otimes B$ is $\in R^{m \times n}$ where $m=m_1 m_2$ and $n=n_1 n_2$. 
Figure \ref{fig:kronecker_product} illustrates the Kronecker product between two small matrices. See \citep{graham2018kronecker} for more detailed information on Kronecker product. 
\subsection{Kronecker factorization}
We can use Kronecker products to represent the weight matrices in Neural Networks (NNs). When decomposing a matrix $W\in R^{m\times n}$, as $A\otimes B$, there are different choices for the shapes of $A$ and $B$.  In fact the dimensions of $A$ i.e $m_1$ and  $n_1$ can be any factor of $m$ and $n$ respectively. The dimensions of $B$ will then be equal to  $m_2=m/m_1$ and $n_2=n/n_1$. We can achieve different compression factors by changing the shape of these two matrices. 

\subsubsection{Memory and computation reduction}
When representing $W$ as $A \otimes B$, the number of elements is reduced from $mn$ to $m_1 n_1+m_2 n_2$. 
Moreover, using the Kronecker product to represent linear layers can reduce the required computation. 
A trivial way to multiply the Kronecker product $A\otimes B$ with an input vector $x$, is to first reconstruct $W$ by obtaining $A\otimes B$  and then multiply the result with $x$, which is extremely inefficient both with respect to memory and computation. A much more efficient way is to use Eq.\ref{Eq:kronecker} which is a well-known property of the Kronecker product that allows obtaining $(A\otimes B)x$ without explicit reconstruction, $A\otimes B$ \citep{lutkepohl1997handbook}:
\begin{equation}
(A\otimes B)x=\mathcal{V}(B \mathcal{R}_{n_2\times n_1}(x) A^\top)  \label{Eq:kronecker}
\end{equation}
where $A^\top$ is $A$ transpose. Here, $\mathcal{V}$ is an operation that transforms a matrix to a vector by stacking its columns and $\mathcal{R}_{n_2\times n_1}(x)$  is an operation that converts a vector $x$ to a matrix of size $n_2 \times n_1$ by dividing the vector to columns of size $n_2$ and concatenating the resulting columns together. The consequence of performing multiplication in this way is that it reduces the number of FLOPs from $(2m_1 m_2-1)n_1 n_2$ to:
\begin{eqnarray}
    \min
    \big((2n_2-1) m_2 n_1+(2n_1-1) m_2 m_1,&\nonumber \\
    (2n_1-1) n_2 m_1+ (2n_2-1)m_2m_1\big)&
    \label{Eq:FLOPs}
\end{eqnarray} 
We use this method for implementation of the Kronecker layers. 
\begin{figure*}[t!]
    \centering
    \includegraphics[trim= 0cm 0cm 1cm 0cm 0cm,clip,width=\textwidth]{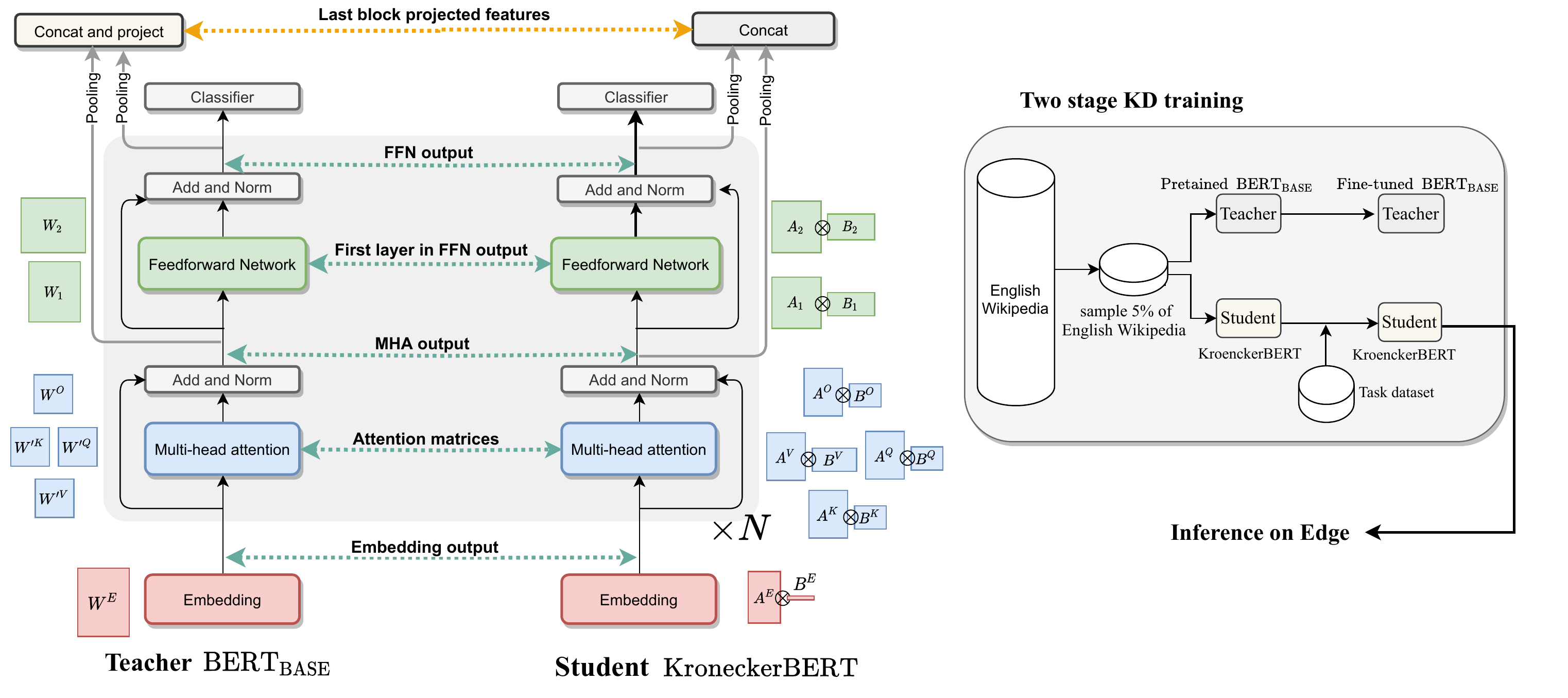}
    \caption{Illustration of the proposed framework. Left: A diagram of the teacher BERT model and the student KronckerBERT.
    Right: The two stage KD methodology used to train KroneckerBERT.}
    \label{fig:KD}
\end{figure*}

\subsection{Kronecker Embedding layer}\label{sec:embedding}
 The embedding layer in large language models is a very large lookup table $X \in R^{v\times d}$, where $v$ is the size of the dictionary and $d$ is the embedding dimension. In order to compress $X$ using Kronecker factorization, the first step is to define the shape of Kronecker factors $A^E$ and $B^E$. We define $A^E$ to be a matrix of size $v\times \frac{d}{n}$ and $B^E$ to be a row vector of size $n$. There are two reasons for defining $B^E$ as a row vector. 1) it allows disentangled embedding of each word since every word has a unique row in $A^E$. 2) the embedding of each word can be obtained efficiently in $\mathcal{O}(d)$. More precisely, the embedding for the $i$'th word in the dictionary can be obtained by the Kronecker product between $A^E_i$ and $B^E$:
 \begin{equation}
     X_i= A^E_i \otimes B^E
 \end{equation}
where$A^E$ is stored as a lookup table. Note that since $A^E_i$ is of size $1\times \frac{d}{n}$ and $B^E$ is of size $1\times n$, the computation complexity of this operation is  $\mathcal{O}(d)$. Figure \ref{fig:embedding} shows an illustration of the Kronecker embedding layer.

\subsection{Kronecker Transformer}\label{sec:transformer}
The Transformer layer is composed of two main components: MHA and FFN. We use Kronecker decomposition to compress both. In the Transformer block, the self-attention mechanism is done by projecting the input into the Key, Query, and Value embeddings and obtaining the attention matrices through the following:
\begin{eqnarray}
\label{Eq:attention}
&O= \frac{QK^T}{\sqrt{d_k}}\\ \nonumber
&\textrm{Attention}(Q,K,V)= \textrm{softmax}(O)V \nonumber
\end{eqnarray}
where $Q$, $K$, and $V$ are obtained by multiplying the input by $W^Q$, $W^K$, $W^V$ respectively.   
In a multi-head attention (MHA) module, there is a separate $W^Q_l$, $W^K_l$, and $W^V_l$ matrix per attention head to allow for a richer representation of the data. In the implementation usually, matrices from all heads are stacked together resulting in 3 matrices. 
\begin{eqnarray}
W'^k=\textrm{concat}( W^K_1,.., W^K_l,... , W^K_L) \\ \nonumber
W'^Q=\textrm{concat}( W^Q_1,.., W^Q_l,... , W^Q_L) \\
W'^V=\textrm{concat}( W^V_1,.., W^V_l,... , W^V_L)  \nonumber
\end{eqnarray}
where $L$ is the number of attention heads. Instead of decomposing the matrices of each head separately, we use Kronecker decomposition after concatenation:
\begin{eqnarray}
    W'^K=A^k\otimes B^K \\ \nonumber
    W'^Q=A^Q\otimes B^Q \\ \nonumber
    W'^V=A^V\otimes B^V   
\end{eqnarray}
By choosing $m_2$ to be smaller than the output dimension of each attention head, matrix $B$ in the Kronecker decomposition is shared among all attention heads resulting in more compression. The result of applying Eq.\ref{Eq:attention} is then fed to a linear mapping ($W^{\textrm{O}}$) to produce the MHA output.
We use Kronecker decomposition for compressing this linear mapping as well the two weight matrices in the subsequent FFN block:
\begin{eqnarray}
W^{\textrm{O}} = A^\textrm{O} \otimes B^\textrm{O}\\
W_1 = A_1 \otimes B_1\\
W_2 = A_2 \otimes B_2
\end{eqnarray}
\raggedbottom
\subsection{Knowledge distillation}
In the following section, we describe how KD is used to improve the training of the KroneckerBERT model.
\begin{table}
\small
\resizebox{\columnwidth}{!}{
\begin{tabular}{| l | c | p{1.5cm}|  p{1.5cm}|  p{0.7cm}|}
\hline
\textbf{Model} & \textbf{CMP} &  $\mathbf{W^K,W^Q,}$ $\mathbf{W^V,W^{O}}$ & $\mathbf{W_1}, [\mathbf{W_2}]^T$ & $\mathbf{W^E}$\\
\hline
& &  \multicolumn{2} {|c|} {${n, m}$} & {${d}$} \\ 
 \hline
BERT$_{\text{BASE}}$ &1$\times$ & 768, 768 & 768, 3072 & 768 \\
\hline 
 &  & \multicolumn{2}{|c|} {${n_1, m_1}$} & {${n}$}  \\  
\hline
KroneckerBERT$_8$& 7.7$\times$ & 384, 384 & 2, 8 & 8 \\
 \hline
 
KroneckerBERT$_{19}$& 19.3$\times$ & 48, 384 & 2, 16  & 12
 \\
 \hline
\end{tabular}
}
\caption{ Configuration of the Kronecker layers for the two KroneckerBERT models used in this paper. CMP stands for compression factor.$n$ and $m$ are respectively the input and output dimensions of the weight matrices ($W\in R^{m\times n}$). $m_1,n_1$ indicates the shape of the first Kronecker factor ($A \in R^{m_1 \times n_1 }$). For embedding layer we only need to set the size of the row vector $B^E \in R^{1\times n}$.}
\label{tab:kronecker_config}
\end{table}
\subsubsection{Intermediate KD}
Let $S$ be the student, and $T$ be the teacher, then for a batch of data $(X,Y)$, we define $f_l^S (X)$ and$ f_l^T (X)$ as the output of the $l^{th}$ layer for the student network and the teacher network respectively. 
The teacher here is the $\text{BERT}_\text{BASE}$ and the student is its corresponding KroneckerBERT that is obtained by replacing the embedding layer and the linear mappings in MHA and FFN modules with Kronecker factors(see Sections \ref{sec:embedding} and \ref{sec:transformer} for details).
Note that like other decomposition methods, when we use Kronecker factorization to compress the model, the number of layers and the dimensions of the input and output of each layer remain intact. Therefore, when performing intermediate layer KD, we can directly obtain the difference in the output of a specific layer in the teacher and student networks without any projection. In the proposed framework, the intermediate KD from the teacher to student occurs at the embedding layer output, attention matrices and FFN outputs:
\begin{eqnarray}
\mathcal{L}_\textrm{Embedding}(X)= \textrm{MSE}\big({E}^S(X),{E}^T(X)\big)\\[6pt]
\mathcal{L}_\textrm{Attention}(X) = \sum_l \textrm{MSE}\big({O}_l^S(X),{O}_l^T(X)\big)
\\
\mathcal{L}_\textrm{FFN}(X) = \sum_l\textrm{MSE}\big({H}_{l}^S(X),{H}_{l}^T(X)\big) 
\end{eqnarray}
where $E^S$ and $E^T$ are the output of the embedding layer from the student and the teacher respectively.
$O_l^S$ and $O_l^T$ are the attention matrices (Eq.\ref{Eq:attention}), ${H}_{l}^S$ and ${H}_{l}^T$ are the outputs of the FFN, of layer $l$ in the student and the teacher respectively.

So far we have been using MSE of the intermediate features for transferring knowledge from the teacher to the student. Therefore, each element in the feature vector of the student and teacher is compared independently of other elements.
Inspired by \citep{sun2020contrastive}, and in order to have a richer comparison between the student and the teacher networks, we also add a projection loss term.
To obtain the projection loss we first average pool the FFN output and attention outputs of the last layer ($H_L$ and $A_L$) and concatenate them together to obtain a feature vector. We project the feature vector obtained from the teacher using a learnable weight matrix $P \in R^{2d\times2d}$, where $d$ is the hidden dimension of the Transformer. We then obtain the projection loss as the MSE between student's features and teacher's projected features:
\begin{eqnarray}
\hspace{-5pt}&&g^S(x)=\textrm{concat}\big[\text{pool}(A^S_L),\text{pool}(H^S_L)\big],\\
\hspace{-5pt}&&g^T(x)=\textrm{concat}\big[\text{pool}(A^T_L),\text{pool}(H^T_L))\big],\\
\hspace{-5pt}&&\mathcal{L}_{Projection}(x)= \textrm{MSE}(g^S(x),P g^T(x)).
\end{eqnarray}
Our final loss is as follows: 
\begin{eqnarray}
\hspace{-5pt}\mathcal{L}(x,y)=&&\sum_{(x,y)}  \mathcal{L}_{Embedding}(x)+\nonumber \\
\hspace{-5pt}&&\mathcal{L}_{Attention}(x) + \mathcal{L}_{FFN}(x)+ \nonumber \\ 
\hspace{-5pt}&&\mathcal{L}_{Projection}(x)+\mathcal{L}_{Logits}(x)+\nonumber \\ 
\hspace{-5pt}&&\mathcal{L}_{CE}(x,y). 
\end{eqnarray}
Note that, unlike some other KD methods where the motivation for this projection is to match the dimension of the student and teacher features, here we use it to obtain a richer representation of the knowledge.
\subsubsection{KD at pre-training}
Inspired by \citep{jiao2019tinybert} we use KD at the pre-training stage to capture the general domain knowledge from the teacher. In pre-training distillation, the teacher is the $\text{BERT}_\text{BASE}$ model that is pre-trained on BookCorpus \citep{zhu2015aligning} and English Wikipedia. Intermediate layer KD is then used to train the KroneckeBERT network in the general domain. KD at pre-training improves the initialization of the Kronecker model for the task-specific KD stage. Similar to \citep{jiao2019tinybert} the loss at pre-training stage only involves the intermediate layers $\mathcal{L}_{Embedding}(x)+\mathcal{L}_{Attention}(x) +\mathcal{L}_{FFN}(x)$.  Unlike \citep{jiao2019tinybert}, we perform pre-training distillation only on a small portion of the dataset (5\% of the English Wikipedia) for a few epochs (3 epochs) which makes our training far more efficient.

\subsection{Model Settings}
The first step of the proposed framework is to design the Kronecker layers by defining the shape of Kronecker factors matrices $A$ and $B$. To do this we need to set the shape of one of these matrices and the other one can be obtained accordingly. Therefore we only searched among different choices for $m_1$ and $n_1$ which are limited to the factors of the original weight matrix (m and n respectively). We used the same configuration for all the matrices in the MHA. Also For the FFN, we chose the configuration for one layer, and for the other layer, the dimensions are swapped. For the embedding layer, since $B^E$ is a row vector, we only need to choose $n$.

The shapes of the Kronecker factors are chosen to obtain the desired compression factor. However, there may be multiple configurations to achieve that, among which we chose the one that leads to minimum FLOPS according to Eq.\ref{Eq:FLOPs}. Table \ref{tab:kronecker_config} summarises the configuration of Kronecker factorization for the two compression factors used in this work.

\subsection{Implementation details}
For KD at the pre-training stage, the KroneckerBERT model is initialized using the teacher (pre-trained BERT$_\text{BASE}$ model). This means that for layers that are not compressed, the values are copied from the teacher to the student. For Kronecker decomposed layers, the  $L_2$ norm approximation \citep{van2000ubiquitous} is used to approximate Kronecker factors ($A$ and $B$) from the pre-trained BERT$_\text{BASE}$ model. 
In the pre-training stage,  5\% of the English Wikipedia was used for 3 epochs. The batch size in pre-training was set to 64 and the learning rate was set e-3. 
After pre-training, the obtained Kronecker model is used to initialize the Kronecker layers in the student model for task-specific fine-tuning. The Prediction layer is initialized from the fine-tuned BERT$_\text{BASE}$ teacher.
For fine-tuning on each task, we optimize the hyper-parameters based on the performance of the model on the dev set. 
See appendix for more details on the results on dev set and the selected hyperparamters.

\begin{table*}
\centering
\small
\setlength\tabcolsep{5pt}
\begin{tabular}{ |l | c | c c c c c c c c |c|}
\hline
\textbf{Model} & \textbf{Params} &   \textbf{MNLI-(m/mm)}  & \textbf{SST-2} & \textbf{MRPC} & \textbf{CoLA} &  \textbf{QQP} &  \textbf{QNLI} & \textbf{RTE} & \textbf{STS-B} & \textbf{Avg} \\

\hline
\hline
BERT$_{\text{BASE}}$  & 108.5M  & 83.9/83.4 &	93.4 &	87.9 &	52.8 &	71.1	& 90.9	& 67	& 85.2	& 79.5 
\\
\hline
BERT$_4$\text{-PKD} & 52.2M & 	{79.9/79.3}	& {89.4} &	82.6 & {24.8}	& 70.2	& 85.1 &  62.3	& 79.8 & 72.6 
\\
TinyBERT & 14.5M & 	\textbf{82.5/81.8}	& \textbf{92.6} &	86.4& \textbf{44.1}	& \textbf{71.3}	&87.7	&  66.6	& 80.4 &  77.0 
\\
LadaBERT$_3$ & 15M &   82.1/81.8 & 89.9 & - & -&  69.4 & 84.5&   - &  -  &  -
\\
KroneckerBERT$_{8}$ & 14.3M & 	\textbf{82.9}/81.7	&  91.2	&  \textbf{88.5}	&  31.2	&  70.8	&  \textbf{88.4}	&  \textbf{66.9} &	\textbf{83.1} 	& 76.1 
\\
\hline

SharedProject & 5.6M & 76.4/75.2 & 84.7 & 84.9 &- & -& -&- & - & - \\
LadaBERT$_4$ & 11M &  75.8/76.1 & 84.0 & - & -&  67.4 &  75.1&   - &  - &-\\
KroneckerBERT$_{19}$ & 5.7M & \textbf{79.4/81.6} &  \textbf{89.2}	& \textbf{86.9}	&  {25.8} 	&  \textbf{69.2}	&  \textbf{86.2} & 62.7 & 78.2	 & 73.1 
\\

\hline
\end{tabular}
\caption{Results on the test set of GLUE official benchmark. The results for BERT, BERT$_4$\text{-PKD} and TinyBERT are taken from \citep{jiao2019tinybert}. For all other baselines the results are taken from their associated papers. SharedProject and LadaBERT refer to \citep{zhao2019extreme} and \citep{mao2020ladabert} respectively. Note that TinyBERT performs pre-training KD on the entire Wikipedia dataset and fine-tuning on Augmented data whereas we only perform pre-training KD on 5\% of the Wikipedia and we do not use data augmentation for fine-tuning.} 
\label{tab:glue_test_results}
\end{table*}
\begin{table}
\small
\begin{tabular}{|p{2.1cm}|p{1.4cm}|p{1.1cm}|p{1.2cm}|}
\hline
\textbf{\space Model} & \textbf{MRPC $\rightarrow$ PAWS
} &  \textbf{RTE $\rightarrow$ HANS} &  
\textbf{SST-2 $\rightarrow$ IMDb} 
 \\
\hline
BERT$_\text{BASE}$ & 61.3 & 50.7&  88.0 
\\
\hline
TinyBERT & 61.3 & 51.2  & 78.5 
\\

KroneckerBERT$_8$ & \textbf{61.4} & \textbf{52.7}& \textbf{81.0} 
\\
\hline
\end{tabular}
\caption{The results of out of distribution experiment. Fined-tuned models on MRPC, RTE and SST-2 are evaluated on PAWS, HANS and IMDb respectively.}
\label{tab:OOD}
\end{table}

\begin{table}
\centering
\setlength\tabcolsep{3pt}
\resizebox{\columnwidth}{!}{
\begin{tabular}{|l|c|c|c|c|c|}
\hline
& & \multicolumn{2}{|c|}{\textbf{SQuAD1.1}} & \multicolumn{2}{|c|}{\textbf{SQuAD2.0}} \\
\textbf{Model} & \textbf{Compress}   &\textbf{EM} &  \textbf{F1} & \textbf{EM} &  \textbf{F1}
 \\
\hline

BERT$_\text{BASE}$ & 1$\times$ & 80.5& 88 & 74.5 & 77.7
\\
\hline
BERT$_4$-PKD & 2.1$\times$ & 70.1& 79.5 & 60.8 & 64.6
\\
TinyBERT & 7.5$\times$ &  72.7 & 82.1  & 68.2 & 71.8
\\
KroneckerBERT$_8$ & 7.6$\times$ &  \textbf{75.4}&   \textbf{84.2} & \textbf{69.9} & \textbf{73.5} \\
\hline
KroneckerBERT$_{19}$ & 19.3$\times$ & 66.7  & 77.8 & 63.0 & 67.0\\

\hline
\end{tabular}

}
\caption{
Results of the baselines and KroneckerBERT
on question SQuAD dev dataset. The results of the baselines are taken from \citep{jiao2019tinybert}. }
\label{tab:squad}
\end{table}

\section{Experiments}
In this section, we compare our KroneckerBERT with the sate-of-the-art compression methods applied to BERT on GLUE and SQuAD. We also investigate the effect of KD through ablation experiments and investigate the robustness of our model to out-of-distribution samples. Moreover we show the speedup results of deploying our model on edge devices and justify our findings.

\subsection{Baselines}
As for baselines we select two categories of compression methods, those with compression factor <10 and those with compression factor >10. For a fair comparison, we select models that have BERT$_\text{BASE}$ as the teacher. In the first category, we both have BERT$_\text{PKD}$ \citep{sun2019patient} with low compression factor, and models with similar compression factor as our KroneckerBERT$_8$: TinyBERT \citep{jiao2019tinybert} and LadaBERT \citep{mao2020ladabert}.

Note that TinyBERT is not directly comparable since they do KD at pre-training on the entire Wikipedia dataset and also they do an extensive augmentation on GLUE for KD in the fine-tuning stage (x20).
We exclude MobileBERT \citep{sun2020mobilebert} since they use a  redesign of BERT$_\text{LARGE}$ with inverted-bottleneck as the teacher.
For the second category, we compare our results with  SharedProject \citep{zhao2019extreme} and \citep{mao2020ladabert} with compression factor in the rage of 10-20. We exclude \citep{zhao2021extremely} since they use BERT$_\text{LARGE}$ as the teacher.
\raggedbottom
\subsection{Results on GLUE benchmark}
We evaluated the proposed framework on the General Language Understanding Evaluation (GLUE) \citep{wang2018glue} benchmark which consists of 9 natural language understanding tasks. We submitted the predictions of our proposed models on the test data sets for different tasks to the official GLUE benchmark (\url{https://gluebenchmark.com/}).
Table \ref{tab:glue_test_results} summarizes the results. The results are divided into two categories: in the upper part of the table KroneckerBERT$_{8}$ are compared to several state-of-the-art KD methods that have a compression factor less than 10. In the lower part of the table, we compare the performance of KroneckerBERT$_{19}$ with state-of-the-art models that have a compression factor greater than 10. We can see that when the compression factor is less than 10, KroneckerBERT outperforms other baselines, except TinyBERT, on every task of GLUE. As for TinyBERT KroneckerBERT has a very similar performance in all the tasks except for CoLA.  Note TinyBERT performs pre-training KD on the entire Wikipedia dataset and uses an extensive data augmentation (20 times the original data) in the fine-tuning stage to obtain these results. Moreover, the average performance of KroneckerBERT$_8$ excluding CoLA is 81.7 compared to 81.2 for TinyBERT. For higher compression factors, KroneckerBERT outperforms all other baselines on all available results.
\subsection{Out of distribution robustness}
It is shown that pre-trained Transformer-based language models are robust to out-of-distribution (OOD) samples \citep{hendrycks2020pretrained}. In this section, we investigate how the proposed compression method affects the OOD robustness of BERT by evaluating the fined-tuned models on MRPC, RTE, and SST-2 on PAWS \citep{zhang2019paws}, HANS \citep{mccoy2019right}, and IMDb \citep{maas-EtAl:2011:ACL-HLT2011} respectively. We compare OOD robustness with the teacher, BERT$_\text{BASE}$ and TinyBERT. TinyBERT fine-tuned checkpoint are obtained from their repsitory. Table \ref{tab:OOD} lists the results. We can see the fine-tuned KroneckerBERT$_8$ models on MRPC and RTE are robust to OOD since there is a small increase in performance compared to BERT$_\text{BASE}$. On IMDb, there is a significant drop in performance after compression, but our KroenckerBERT$_8$ is still  more robust than TinyBERT. In fact, KroencekrBERT$_8$ outperforms TinyBERT on all the three OOD experiments. 
\raggedbottom
\subsection{Results on SQuAD}
In this section, we evaluate the performance of the proposed model on SQuAD datasets.
SQuAD1.1 \citep{2016arXiv160605250R} is a large-scale reading comprehension that contain questions that have answers in given context. SQuAD2.0 \citep{kudo2018sentencepiece} also contains unanswerable questions. 
Table \ref{tab:squad} summarises the performance on dev set. For both SQuAD1.1 and SQuAD2.0, KroneckerBERT$_8$ can significantly outperform other baselines. We have also listed the performance of KroneckerBERT$_{19}$. The results of baselines with higher compression factors on SQuAD were not available.  
\begin{figure}
\includegraphics[trim= 8cm 0cm 0cm 8cm 0cm,clip,width=\columnwidth]{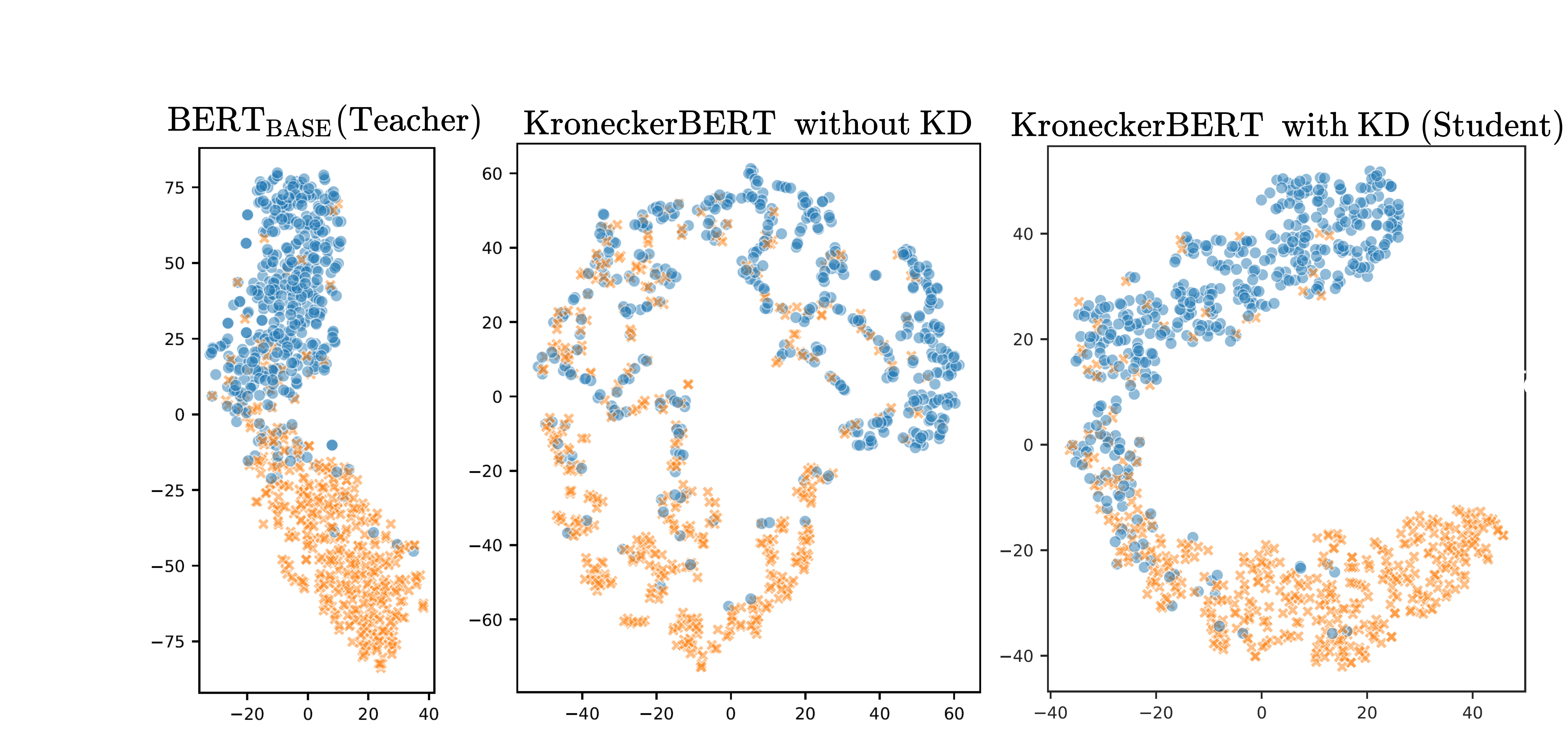}
\caption{T-SNE visualization of the output of the middle Transformer layer of the fine-tuned models on SST-2 dev. Left: Fine-tuned BERT$_\text{BASE}$, middle: KroneckerBERT$_8$  fine-tuned without KD, right: KroneckerBERT$_8$ when trained using KD in two stages. The colours indicate the positive and negative classes.} 
\label{fig:ablation_KD}
\end{figure}
\begin{table}
\centering
\small
\resizebox{\columnwidth}{!}{
\begin{tabular}{| p{1.7cm}| p{1.6cm} | c|c| c|}
\hline  \textbf{Pre-training}  & \textbf{Fine-tuning} & \textbf{MNLI-m} & \textbf{SST-2} & \textbf{MRPC}\\
\hline
 None & No KD & 66.0 & 81.3 & 68.3\\
\hline
 None & KD & 80.7 &  86.2 & 70.3\\
\hline
KD & No KD & 77.0 & 87.2 & 78.17 \\
\hline
KD & KD & \textbf{82.8} & \textbf{90.6} & \textbf{86.6}\\
 \hline
\end{tabular}
}
\caption{Ablation study of the effect of KD in the pre-training and fine-tuning stages on the performance of the model on GLUE dev. The teacher for KD at the pre-training stage and at the fine-tuning stage is the pre-trained and the fined-tuned BERT$_\text{BASE}$  respectively.}
\label{tab:ablation_KD}
\end{table}
\subsection{Ablation study on the effect of KD}
In this section, we investigate the effect of KD in reducing the gap between the compressed KroneckerBERT model and the original BERT$_\text{BASE}$ network. Table \ref{tab:ablation_KD} summarises the results for KroneckerBERT$_8$. Our proposed method uses KD in both the pre-training and the fine-tuning stages. For this ablation study, pre-training is either not performed or is done using KD with the pre-trained BERT$_\text{BASE}$ as the teacher. We perform experiments on 3 tasks from the GLUE benchmark with different sizes of training data, namely MNLI-m, SST-2, and MRPC. We can see that without KD, Kronecker decomposition leads to a significant drop in performance. For MNLI-m we can see that KD at fine-tuning stages is more effective than in performing pre-training with KD. For SST-2 and MRPC, performing pre-training with KD is more effective. For all tasks, the highest performance is obtained when the two-stage KD is used(last row). 

We also used t-SNE to visualize the output of the FFN of the middle layer (layer 6) of the fine-tuned KroneckerBERT$_8$ with and without KD in comparison with the fine-tuned teacher, on SST-2 dev. Figure \ref{fig:ablation_KD} shows the results. See how KD helps the features of the middle layer to be more separable with respect to the task compared to the no KD case.

\begin{table}
\centering
\setlength\tabcolsep{3pt}
\resizebox{\columnwidth}{!}{
\begin{tabular}{|l|c|c|c|c|}
\hline
\textbf{Model} & \textbf{FLOPS} & 
\multicolumn{3}{|c|} {\textbf{Inference Speedup}} \\
\hline
& & Smartphone 1& Smartphone 2 &  CPU \\ 
\hline
BERT$_\text{BASE}$ &
22B& 1 & 1 & 1  \\
\hline
KroneckerBERT$_8$ & 5.5B & 0.94 & 0.74 & 0.81 \\
KroneckerBERT$_{19}$ & 1.4B & 2.4 & 1.93 & 1.1 \\
\hline
\end{tabular}
}
\caption{Number of FLOPS and inference speed up on 2 smartphones and Intel CPU (Intel Xeon Gold6140) for the proposed model with respect to BERT$_\text{BASE}$.} 
\label{tab:latancy}
\end{table}
\subsection{Inference speedup: Discussion and results}
Decomposition methods like SVD, Tucker, etc. all try to represent a given tensor as multiple significantly smaller tensors and thus reduce the number of parameters. The same is also true for the number of operations involved in the forward pass: instead of one large matrix operation, we often have multiple small matrix operations which can lead to a reduction in the number of FLOPS. However, this reduction in the number of FLOPS may not lead to latency reduction since these smaller matrix operations have to be performed in serial. Kronecker decomposition is no exception in this regard. In order to perform Kronecker product matrix multiplication $(A \otimes B)X$ we need to multiply 3 matrices as in Eq.\ref{Eq:kronecker}. Multiplication of 3 matrices is done through two matrix-matrix multiplication in serial. Serial nature of this operation limits utilization of parallel processing units on modern CPUs and GPUs. Therefore without modification of the underlying software/hardware, the reduction of the number of FLOPs may not lead to latency reduction on high-performance devices. 

In order to investigate how this reduction in the number of FLOPS is translated to a reduction in the latency, for high compression factors, we serialized the Kronecker$_{19}$ using  PyTorch's TorchScript and deployed it on two smartphones as well as on a high-performance CPU (18 core Intel\copyright Xeon\copyright GOLD 6140). The specification of octa-core processors for smartphone 1 and 2 is (4x2.4GHz, 4x1.8GHz) and (2x2.6 GHz,2x1.9GHz, 4x1.8GHz) respectively.
We measured the latency over 50 runs of a single input sequence of length 128. The results are listed in Table \ref{tab:latancy}. We can see that for the high compression factor of 19, the compressed model can have up to 2.4 speed up on edge devices. However, on the high-performance CPU with much higher memory and computation capacity using KroneckerBERT does not lead to speed-up which confirms the above argument.

\section{Conclusion}
We introduced a novel framework for compressing Transformer-based language models. The proposed model uses Kronecker decomposition for the compression of the embedding layer and the linear mappings within the Transformer blocks. The proposed framework was used to compress the BERT$_\text{BASE}$ model. We used an efficient two-stage KD method to train the compressed model. We show that the proposed framework can significantly reduce the size and the number of computations while outperforming state-of-the-art for high compression factors of at least $10\times$. The proposed methodology can be applied to any other Transformer based language model.

\section*{Acknowledgements}
Authors would like to thank Mahdi Zolnouri for deploying the models on cellphones and obtaining latency results and Aref Jaffari for preparing the OOD datasets. We also would like to thank Seyed Alireza Ghaffari and Eyy\"ub Sari for informative discussions throughout this project.

\bibliography{references}

\begin{thebibliography}{36}
\expandafter\ifx\csname natexlab\endcsname\relax\def\natexlab#1{#1}\fi

\bibitem[{Buciluǎ et~al.(2006)Buciluǎ, Caruana, and
  Niculescu-Mizil}]{bucilu2006model}
Cristian Buciluǎ, Rich Caruana, and Alexandru Niculescu-Mizil. 2006.
\newblock Model compression.
\newblock In \emph{Proceedings of the 12th ACM SIGKDD international conference
  on Knowledge discovery and data mining}, pages 535--541.

\bibitem[{Devlin et~al.(2018)Devlin, Chang, Lee, and
  Toutanova}]{devlin2018bert}
Jacob Devlin, Ming-Wei Chang, Kenton Lee, and Kristina Toutanova. 2018.
\newblock Bert: Pre-training of deep bidirectional transformers for language
  understanding.
\newblock \emph{arXiv preprint arXiv:1810.04805}.

\bibitem[{Gong et~al.(2014)Gong, Liu, Yang, and Bourdev}]{gong2014compressing}
Yunchao Gong, Liu Liu, Ming Yang, and Lubomir Bourdev. 2014.
\newblock Compressing deep convolutional networks using vector quantization.
\newblock \emph{arXiv preprint arXiv:1412.6115}.

\bibitem[{Graham(2018)}]{graham2018kronecker}
Alexander Graham. 2018.
\newblock \emph{Kronecker products and matrix calculus with applications}.
\newblock Courier Dover Publications.

\bibitem[{Han et~al.(2015)Han, Mao, and Dally}]{han2015deep}
Song Han, Huizi Mao, and William~J Dally. 2015.
\newblock Deep compression: Compressing deep neural networks with pruning,
  trained quantization and huffman coding.
\newblock \emph{arXiv preprint arXiv:1510.00149}.

\bibitem[{Hendrycks et~al.(2020)Hendrycks, Liu, Wallace, Dziedzic, Krishnan,
  and Song}]{hendrycks2020pretrained}
Dan Hendrycks, Xiaoyuan Liu, Eric Wallace, Adam Dziedzic, Rishabh Krishnan, and
  Dawn Song. 2020.
\newblock Pretrained transformers improve out-of-distribution robustness.
\newblock \emph{arXiv preprint arXiv:2004.06100}.

\bibitem[{Hinton et~al.(2015)Hinton, Vinyals, and Dean}]{hinton2015distilling}
Geoffrey Hinton, Oriol Vinyals, and Jeff Dean. 2015.
\newblock Distilling the knowledge in a neural network.
\newblock \emph{arXiv preprint arXiv:1503.02531}.

\bibitem[{Jiao et~al.(2019)Jiao, Yin, Shang, Jiang, Chen, Li, Wang, and
  Liu}]{jiao2019tinybert}
Xiaoqi Jiao, Yichun Yin, Lifeng Shang, Xin Jiang, Xiao Chen, Linlin Li, Fang
  Wang, and Qun Liu. 2019.
\newblock Tinybert: Distilling bert for natural language understanding.
\newblock \emph{arXiv preprint arXiv:1909.10351}.

\bibitem[{Khrulkov et~al.(2019)Khrulkov, Hrinchuk, Mirvakhabova, and
  Oseledets}]{khrulkov2019tensorized}
Valentin Khrulkov, Oleksii Hrinchuk, Leyla Mirvakhabova, and Ivan Oseledets.
  2019.
\newblock Tensorized embedding layers for efficient model compression.
\newblock \emph{arXiv preprint arXiv:1901.10787}.

\bibitem[{Kim and Awadalla(2020)}]{kim2020fastformers}
Young~Jin Kim and Hany~Hassan Awadalla. 2020.
\newblock Fastformers: Highly efficient transformer models for natural language
  understanding.
\newblock \emph{arXiv preprint arXiv:2010.13382}.

\bibitem[{Kudo and Richardson(2018)}]{kudo2018sentencepiece}
Taku Kudo and John Richardson. 2018.
\newblock Sentencepiece: A simple and language independent subword tokenizer
  and detokenizer for neural text processing.
\newblock \emph{arXiv preprint arXiv:1808.06226}.

\bibitem[{Li et~al.(2018)Li, Kulhanek, Wang, Zhao, and Wu}]{li2018slim}
Zhongliang Li, Raymond Kulhanek, Shaojun Wang, Yunxin Zhao, and Shuang Wu.
  2018.
\newblock Slim embedding layers for recurrent neural language models.
\newblock In \emph{Proceedings of the AAAI Conference on Artificial
  Intelligence}, volume~32.

\bibitem[{Lioutas et~al.(2020)Lioutas, Rashid, Kumar, Haidar, and
  Rezagholizadeh}]{lioutas2020improving}
Vasileios Lioutas, Ahmad Rashid, Krtin Kumar, Md~Akmal Haidar, and Mehdi
  Rezagholizadeh. 2020.
\newblock Improving word embedding factorization for compression using
  distilled nonlinear neural decomposition.
\newblock In \emph{Proceedings of the 2020 Conference on Empirical Methods in
  Natural Language Processing: Findings}, pages 2774--2784.

\bibitem[{Lutkepohl(1997)}]{lutkepohl1997handbook}
Helmut Lutkepohl. 1997.
\newblock Handbook of matrices.
\newblock \emph{Computational statistics and Data analysis}, 2(25):243.

\bibitem[{Maas et~al.(2011)Maas, Daly, Pham, Huang, Ng, and
  Potts}]{maas-EtAl:2011:ACL-HLT2011}
Andrew~L. Maas, Raymond~E. Daly, Peter~T. Pham, Dan Huang, Andrew~Y. Ng, and
  Christopher Potts. 2011.
\newblock \href {http://www.aclweb.org/anthology/P11-1015} {Learning word
  vectors for sentiment analysis}.
\newblock In \emph{Proceedings of the 49th Annual Meeting of the Association
  for Computational Linguistics: Human Language Technologies}, pages 142--150,
  Portland, Oregon, USA. Association for Computational Linguistics.

\bibitem[{Mao et~al.(2020)Mao, Wang, Wu, Zhang, Wang, Yang, Zhang, Tong, and
  Bai}]{mao2020ladabert}
Yihuan Mao, Yujing Wang, Chufan Wu, Chen Zhang, Yang Wang, Yaming Yang, Quanlu
  Zhang, Yunhai Tong, and Jing Bai. 2020.
\newblock Ladabert: Lightweight adaptation of bert through hybrid model
  compression.
\newblock \emph{arXiv preprint arXiv:2004.04124}.

\bibitem[{McCoy et~al.(2019)McCoy, Pavlick, and Linzen}]{mccoy2019right}
R~Thomas McCoy, Ellie Pavlick, and Tal Linzen. 2019.
\newblock Right for the wrong reasons: Diagnosing syntactic heuristics in
  natural language inference.
\newblock \emph{arXiv preprint arXiv:1902.01007}.

\bibitem[{Prato et~al.(2019)Prato, Charlaix, and
  Rezagholizadeh}]{prato2019fully}
Gabriele Prato, Ella Charlaix, and Mehdi Rezagholizadeh. 2019.
\newblock Fully quantized transformer for machine translation.
\newblock \emph{arXiv preprint arXiv:1910.10485}.

\bibitem[{Radford et~al.(2019)Radford, Wu, Child, Luan, Amodei, and
  Sutskever}]{radford2019language}
Alec Radford, Jeffrey Wu, Rewon Child, David Luan, Dario Amodei, and Ilya
  Sutskever. 2019.
\newblock Language models are unsupervised multitask learners.
\newblock \emph{OpenAI blog}, 1(8):9.

\bibitem[{{Rajpurkar} et~al.(2016){Rajpurkar}, {Zhang}, {Lopyrev}, and
  {Liang}}]{2016arXiv160605250R}
Pranav {Rajpurkar}, Jian {Zhang}, Konstantin {Lopyrev}, and Percy {Liang}.
  2016.
\newblock \href {http://arxiv.org/abs/1606.05250} {{SQuAD: 100,000+ Questions
  for Machine Comprehension of Text}}.
\newblock \emph{arXiv e-prints}, page arXiv:1606.05250.

\bibitem[{Sanh et~al.(2019)Sanh, Debut, Chaumond, and
  Wolf}]{sanh2019distilbert}
Victor Sanh, Lysandre Debut, Julien Chaumond, and Thomas Wolf. 2019.
\newblock Distilbert, a distilled version of bert: smaller, faster, cheaper and
  lighter.
\newblock \emph{arXiv preprint arXiv:1910.01108}.

\bibitem[{Shoeybi et~al.(2019)Shoeybi, Patwary, Puri, LeGresley, Casper, and
  Catanzaro}]{shoeybi2019megatron}
Mohammad Shoeybi, Mostofa Patwary, Raul Puri, Patrick LeGresley, Jared Casper,
  and Bryan Catanzaro. 2019.
\newblock Megatron-lm: Training multi-billion parameter language models using
  model parallelism.
\newblock \emph{arXiv preprint arXiv:1909.08053}.

\bibitem[{Sun et~al.(2019)Sun, Cheng, Gan, and Liu}]{sun2019patient}
Siqi Sun, Yu~Cheng, Zhe Gan, and Jingjing Liu. 2019.
\newblock Patient knowledge distillation for bert model compression.
\newblock \emph{arXiv preprint arXiv:1908.09355}.

\bibitem[{Sun et~al.(2020{\natexlab{a}})Sun, Gan, Cheng, Fang, Wang, and
  Liu}]{sun2020contrastive}
Siqi Sun, Zhe Gan, Yu~Cheng, Yuwei Fang, Shuohang Wang, and Jingjing Liu.
  2020{\natexlab{a}}.
\newblock Contrastive distillation on intermediate representations for language
  model compression.
\newblock \emph{arXiv preprint arXiv:2009.14167}.

\bibitem[{Sun et~al.(2020{\natexlab{b}})Sun, Yu, Song, Liu, Yang, and
  Zhou}]{sun2020mobilebert}
Zhiqing Sun, Hongkun Yu, Xiaodan Song, Renjie Liu, Yiming Yang, and Denny Zhou.
  2020{\natexlab{b}}.
\newblock Mobilebert: a compact task-agnostic bert for resource-limited
  devices.
\newblock \emph{arXiv preprint arXiv:2004.02984}.

\bibitem[{Thakker et~al.(2020)Thakker, Whatamough, Mattina, and
  Beu}]{thakker2020compressing}
Urmish Thakker, Paul Whatamough, Matthew Mattina, and Jesse Beu. 2020.
\newblock Compressing language models using doped kronecker products.
\newblock \emph{arXiv preprint arXiv:2001.08896}.

\bibitem[{Van~Loan(2000)}]{van2000ubiquitous}
Charles~F Van~Loan. 2000.
\newblock The ubiquitous kronecker product.
\newblock \emph{Journal of computational and applied mathematics},
  123(1-2):85--100.

\bibitem[{Wang et~al.(2018)Wang, Singh, Michael, Hill, Levy, and
  Bowman}]{wang2018glue}
Alex Wang, Amanpreet Singh, Julian Michael, Felix Hill, Omer Levy, and Samuel~R
  Bowman. 2018.
\newblock Glue: A multi-task benchmark and analysis platform for natural
  language understanding.
\newblock \emph{arXiv preprint arXiv:1804.07461}.

\bibitem[{Xu et~al.(2020)Xu, Zhou, Ge, Wei, and Zhou}]{xu2020bertoftheseus}
Canwen Xu, Wangchunshu Zhou, Tao Ge, Furu Wei, and Ming Zhou. 2020.
\newblock \href {http://arxiv.org/abs/2002.02925} {Bert-of-theseus: Compressing
  bert by progressive module replacing}.

\bibitem[{Yang et~al.(2019)Yang, Dai, Yang, Carbonell, Salakhutdinov, and
  Le}]{yang2019xlnet}
Zhilin Yang, Zihang Dai, Yiming Yang, Jaime Carbonell, Ruslan Salakhutdinov,
  and Quoc~V Le. 2019.
\newblock Xlnet: Generalized autoregressive pretraining for language
  understanding.
\newblock \emph{arXiv preprint arXiv:1906.08237}.

\bibitem[{Yu et~al.(2017)Yu, Liu, Wang, and Tao}]{yu2017compressing}
Xiyu Yu, Tongliang Liu, Xinchao Wang, and Dacheng Tao. 2017.
\newblock On compressing deep models by low rank and sparse decomposition.
\newblock In \emph{Proceedings of the IEEE Conference on Computer Vision and
  Pattern Recognition}, pages 7370--7379.

\bibitem[{Zhang et~al.(2019)Zhang, Baldridge, and He}]{zhang2019paws}
Yuan Zhang, Jason Baldridge, and Luheng He. 2019.
\newblock Paws: Paraphrase adversaries from word scrambling.
\newblock \emph{arXiv preprint arXiv:1904.01130}.

\bibitem[{Zhao et~al.(2019)Zhao, Gupta, Song, and Zhou}]{zhao2019extreme}
Sanqiang Zhao, Raghav Gupta, Yang Song, and Denny Zhou. 2019.
\newblock Extreme language model compression with optimal subwords and shared
  projections.
\newblock \emph{arXiv preprint arXiv:1909.11687}.

\bibitem[{Zhao et~al.(2021)Zhao, Gupta, Song, and Zhou}]{zhao2021extremely}
Sanqiang Zhao, Raghav Gupta, Yang Song, and Denny Zhou. 2021.
\newblock Extremely small bert models from mixed-vocabulary training.
\newblock In \emph{Proceedings of the 16th Conference of the European Chapter
  of the Association for Computational Linguistics: Main Volume}, pages
  2753--2759.

\bibitem[{Zhou and Wu(2015)}]{zhou2015compression}
Shuchang Zhou and Jia-Nan Wu. 2015.
\newblock Compression of fully-connected layer in neural network by kronecker
  product.
\newblock \emph{arXiv preprint arXiv:1507.05775}.

\bibitem[{Zhu et~al.(2015)Zhu, Kiros, Zemel, Salakhutdinov, Urtasun, Torralba,
  and Fidler}]{zhu2015aligning}
Yukun Zhu, Ryan Kiros, Rich Zemel, Ruslan Salakhutdinov, Raquel Urtasun,
  Antonio Torralba, and Sanja Fidler. 2015.
\newblock Aligning books and movies: Towards story-like visual explanations by
  watching movies and reading books.
\newblock In \emph{Proceedings of the IEEE international conference on computer
  vision}, pages 19--27.

\end{thebibliography}
\bibliographystyle{acl_natbib}

\appendix

\end{document}